\documentclass{optica-article}

\journal{opticajournal}

\articletype{Research Article}

\usepackage{multirow}
\newcommand\ie{\textit{i.e.}}
\newcommand\eg{\textit{e.g.}}

\begin{document}

\title{Noise2Image: Noise-Enabled Static Scene Recovery for Event Cameras}

\author{Ruiming Cao,\authormark{1, \dag, *} Dekel Galor,\authormark{2, \dag, *} Amit Kohli,\authormark{2} Jacob L Yates,\authormark{3} and Laura Waller\authormark{2}}

\address{\authormark{1}Department of Bioengineering, University of California, Berkeley, Berkeley, CA 94720, USA\\
\authormark{2}Department of Electrical Engineering and Computer Sciences, University of California, Berkeley, Berkeley, CA 94720, USA\\
\authormark{3}Department of Vision Science, University of California, Berkeley, Berkeley, CA 94720, USA\\
\authormark{\dag}The authors contributed equally to this work.\\
}

\email{\authormark{*}rcao@berkeley.edu, galor@berkeley.edu}

\begin{abstract*} 
Event cameras, also known as dynamic vision sensors, are an emerging modality for measuring fast dynamics asynchronously. 
Event cameras capture changes of log-intensity over time as a stream of `events' and generally cannot measure intensity itself; hence, they are only used for imaging dynamic scenes. However, fluctuations due to random photon arrival inevitably trigger \emph{noise events}, even for static scenes. While previous efforts have been focused on filtering out these undesirable noise events to improve signal quality, we find that, in the photon-noise regime, these noise events are correlated with the static scene intensity. We analyze the noise event generation and model its relationship to illuminance. Based on this understanding, we propose a method, called Noise2Image, to leverage the illuminance-dependent noise characteristics to recover the static parts of a scene, which are otherwise invisible to event cameras. We experimentally collect a dataset of noise events on static scenes to train and validate Noise2Image. Our results provide a novel approach for capturing static scenes in event cameras, solely from noise events, without additional hardware.
\end{abstract*}

\section{Introduction}
Event cameras~\cite{mahowald1994silicon}, also known as neuromorphic cameras or dynamic vision sensors, are an emerging modality for capturing dynamic scenes. Their ability to capture data at a much faster rate than conventional cameras has led to applications in high-speed navigation, augmented reality, and real-time 3D reconstruction~\cite{gallego2020event}. Unlike a conventional CMOS camera that outputs intensity images at fixed time intervals, an event camera detects brightness changes at each pixel asynchronously. When the brightness change at a pixel exceeds some threshold, an event is recorded. The output from an event consists of three elements: a timestamp, the spatial coordinate of the triggered pixel, and a binary polarity (indicating whether it is an increase or decrease of brightness). 

While this scheme allows event cameras to operate beyond conventional framerates, it makes them blind to the static components of a scene, which induce no brightness changes over time. This issue is especially prevalent when the camera is not moving and therefore provides no information about the static background. Even though event cameras are not designed to capture an intensity image, it is often needed for downstream applications such as initializing motion tracking algorithms~\cite{angelopoulos2020event}. To mitigate this issue, some event cameras include a conventional frame-based sensor in the pixel circuit to simultaneously image both events and traditional intensity images~\cite{brandli2014real, berner2013240}. Alternatively, a frame-based camera can be installed in parallel, using either a beam splitter~\cite{hidalgo2022event, zou2021learning, zhu2021neuspike} or additional view registration~\cite{wang2021asynchronous}. Another approach uses a built an opto-mechanical device that rotates a wedge prism to generate events on static scenes~\cite{he2024microsaccade}.  All of these solutions introduce additional hardware and increase cost, complexity, size, and power consumption, highlighting the challenge of capturing static scenes without adding significant overhead.

\begin{figure}[t] 
\centering
\includegraphics[width=\textwidth]{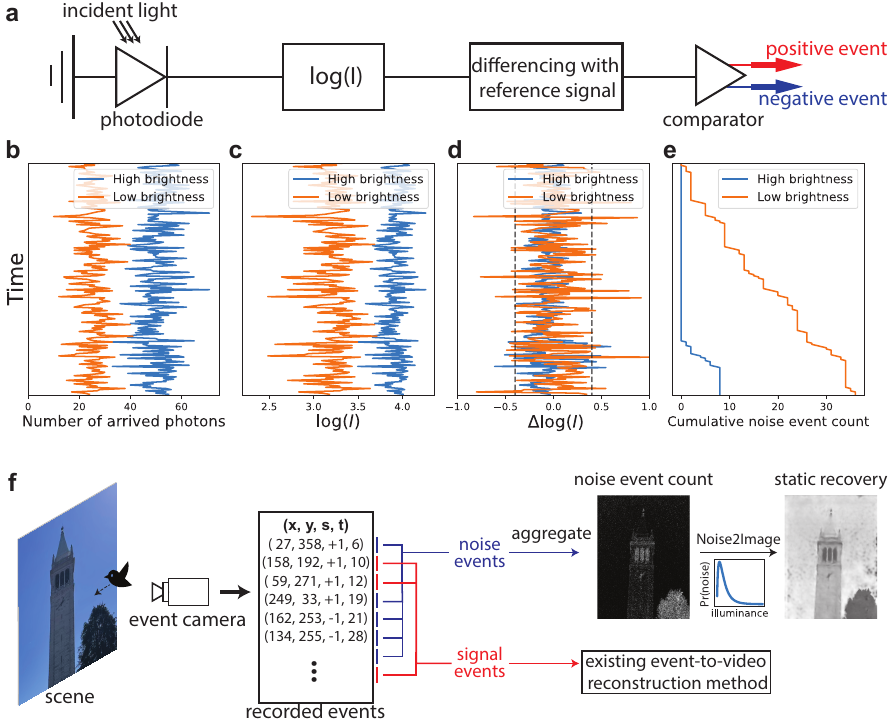}
\caption{Overview of Noise2Image: \textbf{a}. An event pixel circuit compares the change of logarithmic light intensity and produces an event when the change passes the contrast threshold. \textbf{b-e}. To illustrate the illuminance-dependency of noise events on static scenes, we simulate the photon arrival process for low or high brightness scenes. The variance of logarithmic intensity is higher for the low brightness scene, and thus its changes from the reference signal triggering the previous event is also more dramatic, resulting much noise events under low brightness. The dotted black lines in \textbf{d} indicate the contrast thresholds for positive and negative events. 
\textbf{f}. Noise2Image method only takes noise events to reconstruct the static scene intensity. Noise2Image relies on characterizing the relationship between noise events and using learned priors to resolve ambiguities. For scenes with both static and dynamic components, signal events triggered by intensity changes can be fed into existing event-to-video reconstruction methods. 
} \label{fig-pipeline}
\end{figure}

Our work leverages the fact that, even when the scene is static, event cameras still produce noise events. We focus on low and moderate-brightness regimes (\eg, room light, outdoor sunset)~\cite{guo2022low, gracca2023shining, graca2021unraveling} in which the dominant source of noise is photon noise --- random fluctuations in the photon arrival process. In contrast, the high-brightness regime (\eg, outdoor daylight) includes leakage noise events~\cite{nozaki2017temperature}, which we do not model here. While photon noise is well-studied in the context of frame-based scenes, the events triggered by photon noise are commonly treated as general background noise activity to be filtered out. 

In this work we propose a method called Noise2Image that reconstructs a static scene from its event noise statistics, with no hardware modifications. First, we derive a statistical noise model describing how noise event generation correlates with scene intensity, which shows a good correspondence with our experimental measurements. Unlike in conventional sensors, where photon noise grows with the signal, we find that for event cameras, the number of events triggered by photon noise is mostly negatively correlated with the illuminance level due to the logarithmic sensitivity of the sensor. Imaging the static scene then amounts to inverting this intensity-to-noise process. However, the mapping is one-to-many, so not directly invertible; thus, we rely on a learned prior to resolve ambiguities. To train and validate our method, we experimentally collect a dataset of event recordings on static scenes. We demonstrate that Noise2Image can recover photo-realistic images from noise events alone, or can be used to recover the static parts of scenes with dynamics. In addition to testing on in-distribution data, we demonstrate the robustness of this approach with out-of-distribution testing data and live scenes.

\section{Related Work}

\subsection{Event-to-video reconstruction}
Event-to-video reconstruction (E2VID) is a class of methods that recover high frame-rate video from event recordings. These methods can capture high-speed dynamics without motion blur. Because event cameras only capture scene changes, many E2VID approaches include one or two traditional frame-based images, using events to deblur~\cite{scheerlinck2018continuous, jiang2020learning, pan2019bringing, wang2020event}, synthesize adjacent frames~\cite{wang2019event}, or interpolate temporally~\cite{tulyakov2021time, tulyakov2022time, zhang2022unifying, wang2023event}.

In contrast, a more challenging class of E2VID reconstruction only uses events for video reconstruction. Because the initial scene intensity is unknown and only the relative intensity changes are measured, reconstruction requires either explicit modeling of spatiotemporal relationships~\cite{bardow2016simultaneous} or deep neural networks as a data prior to fill in the missing information~\cite{rebecq2019high, stoffregen2020reducing, Scheerlinck20wacv, weng2021event, cadena2021spade, paredes2021back, ercan2023hypere2vid}. Because it is difficult to collect event recordings paired with the corresponding frame-based videos at scale, the data prior is obtained through synthetic data generation~\cite{rebecq2019high, stoffregen2020reducing} using an event camera simulator~\cite{rebecq2018esim}. While E2VID methods are effective for recordings with object motion or camera motion, the reconstruction of static scenes is still out of reach since no events should be triggered without motion. Our Noise2Image method can be considered complementary to E2VID; one might use Noise2Image to recover the static parts of the scene and E2VID to recover the dynamic parts.

\subsection{Event camera noise characterization}
An event recording often contains a number of events not associated with intensity changes, which are termed noise events or background activity~\cite{delbruck1995analog, gracca2023shining}. Noise events are attributed to two main sources: photon noise and leakage current~\cite{nozaki2017temperature, graca2021unraveling, guo2022low}. 
Photon noise is the dominant source of noise in low-brightness conditions~\cite{Lichtsteiner2008A11, graca2021unraveling}, while leakage noise events dominate in high-brightness conditions~\cite{nozaki2017temperature, graca2021unraveling}. 
Although the noise model of event cameras is less studied than traditional sensors, in the lower-light regime, it is generally believed that noise events become less likely to trigger as intensity increases~\cite{ding2023mlb,hu2021v2e}. In event camera simulators, the events triggered by photon noise are modeled as a Poisson process, with the noise event rate linearly decreasing with intensity~\cite{hu2021v2e}. The intensity dependency of noise events has been experimentally measured in~\cite{graca2021unraveling}, and shows a non-monotonic relationship with intensity in low light. Our experimental results are consistent with this trend, and we derive a theoretical noise model which explains the relationship between intensity and noise events.

\subsection{Event denoising}
While denoising methods for CMOS or CCD sensors often focus on building an accurate noise model, event camera denoising methods instead emphasize noise detection by identifying the ``signal'' events corresponding to changes in the scene and removing everything else. Because natural scene changes are inherently spatiotemporal, an event triggered by real signal should be accompanied by a number of other events at the neighboring spatial and temporal locations. With this observation, a background activity filter (BAF) (also called nearest neighbor filter) is used to identify real events by checking the time difference between a new event and the most recent previous event in its proximate spatial location and rejecting events when the time difference passes a threshold~\cite{delbruck2008frame, czech2016evaluating, padala2018noise}. Similarly, the noise events can also be found through spatiotemporal local plane fitting~\cite{benosman2013event}. BAF is simple to compute and can be implemented in hardware with limited computational resources~\cite{liu2015design, barrios2018less}. Guo and Delbruck further extended BAF by reducing its memory footprint and incorporating structural information from a data-driven signal-versus-noise classifier~\cite{guo2022low}. A recent study points out that events are often triggered simultaneously by noise and signal and therefore treats denoising as a regression problem to predict the noise likelihood~\cite{Baldwin_2020_CVPR}. Here, we assume access to a good event denoiser~\cite{feng2020event, ding2023mlb} that isolates noise events resulting from static components of a scene.

\subsection{Event camera datasets}
Unlike frame-based cameras, the collection and curation of event camera datasets is challenging, largely due to the inaccessible hardware and the scarcity of online data repositories (\eg, flickr, Google Images for framed-based images). Nevertheless, since large datasets are critical for machine learning~\cite{krizhevsky2012imagenet, dosovitskiy2020image}, there are a handful of recent efforts to systematically generate large-scale event camera datasets. One common approach is to display images on a monitor and record them using an event camera~\cite{serrano2015poker, orchard2015converting, li2017cifar10, kim2021n}. The motion in the scene can then be generated by either moving the displayed image on the monitor~\cite{serrano2015poker, li2017cifar10} or moving the camera~\cite{orchard2015converting, kim2021n}. Alternatively, event datasets can also be generated from frame-based video datasets by over-sampling in time and feeding into an event camera simulator~\cite{gehrig2020video, hu2021v2e, zhu2021eventgan}. In this work, we generate an NE2I dataset using experimental measurements and synthetic noise, described in Sec.~\ref{sec-theory}.

\section{Modeling Noise Event Statistics}
\label{sec-theory}

First, we develop a model of noise event statistics for later use in our synthetic dataset generation and image reconstruction steps. An event $e$ is triggered when a pixel $\left(x, y\right)$ detects a change in logarithmic intensity, $\log(I)$, greater than the contrast threshold, $\epsilon$, at the time $t$. The polarity $s = +1$ if the change is positive, and $s = -1$ if the change is negative. The triggering condition for an event can be written as 
\begin{equation}
\left( \log(I(x, y, t)) - \log(I(x, y, t_0)) \right) \cdot s > \epsilon,
\end{equation}
\noindent where $t_0$ is the timestamp for the most recent event at the same pixel. The logarithmic sensitivity ensures that this triggering condition is adaptive to the brightness level, \eg, a tenfold increase from 1 lux to 10 lux would result in a similar response as a tenfold increase from 100 lux to 1000 lux. As pixels operate asynchronously, we will proceed by considering events at an arbitrary pixel and drop the $\left(x, y\right)$ notation for simplicity.

By design, there should be no events triggered by a static scene, because there will be no changes in intensity. However, the inherent randomness in the photon arrival process means that there will be fluctuations in the detected intensity and thus noise events triggered. Consider a static scene in which a pixel sees $n$ photons over a short period of time. $n$ follows a Poisson distribution with the average photon count $\lambda \propto I$, which can be further approximated as a Gaussian distribution for the light levels we work with in photography conditions ($\lambda > 10$). We similarly write the photon count during the last event trigger as $n_0$, and both $n, n_0 \overset{\text{i.i.d.}}{\sim} \mathcal{N}\left( \lambda, \lambda \right)$. 

\begin{figure}[bht]
\centering
\includegraphics[width=0.6 \textwidth]{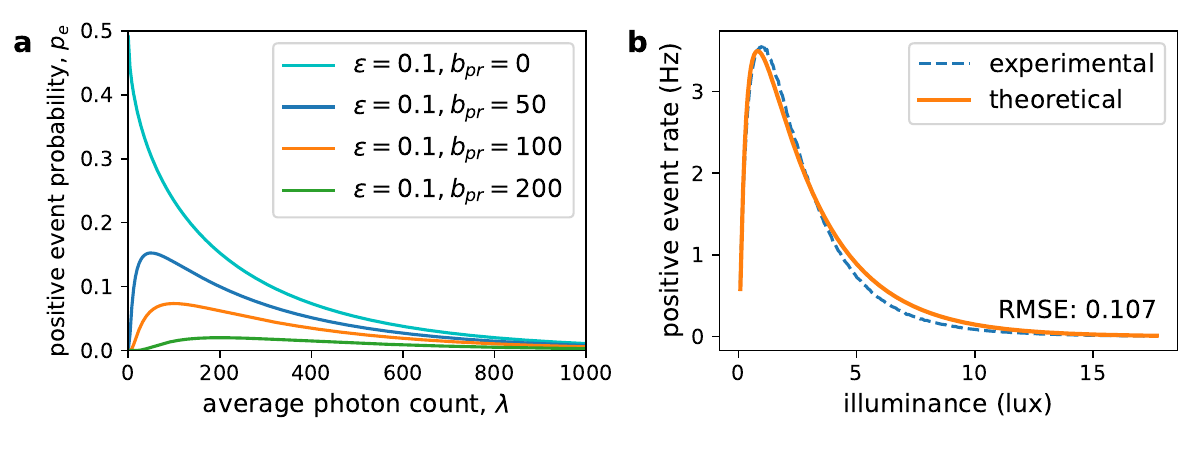}  
\caption{\textbf{a}. Theoretical noise event probability, $p_e$, versus the average photon count, $\lambda$, at different photoreceptor bias values, $b_{pr}$ ($\epsilon$ is the contrast threshold). \textbf{b}. Experimentally measured noise event rate versus illuminance (proportional to $\lambda$) matches well with our theoretical model after fitting parameters $\epsilon, b_{pr}$ and $N$.}  
\label{fig-theory-curve}
\end{figure}

To trigger a positive event, we need $\log\left( n\right) - \log\left( n_0\right) > \epsilon$, which can be re-written as $n - n_0 e^\epsilon > 0$. Under the Gaussian approximation, $n - n_0 e^\epsilon \sim \mathcal{N} \left(\lambda\left( 1 - e^\epsilon \right), \lambda \left( 1 + e^{2\epsilon} \right) \right) $. Thus, we can write the probability of triggering a positive event,
\begin{equation}
Pr\left( n - n_0 e^\epsilon > 0 \right) = \frac{1}{2} - \frac{1}{2} \textit{erf} \left( \frac{\lambda \left( e^\epsilon - 1\right) }{\sqrt{2 \lambda \left( 1 + e^{2\epsilon} \right)}} \right),
\end{equation}
where $\textit{erf}$ denotes the error function. The derivation for negative events is similar. This function is plotted in Fig.~\ref{fig-theory-curve}a (cyan curve) and reveals a monotonically decreasing trend between noise event probability and illuminance. In other words, as the illuminance increases, it is less likely to find $n, n_0$ that satisfies the triggering condition. This trend matches with previous literature~\cite{ding2023mlb, graca2021unraveling, gracca2023shining}.

For low-light conditions, where $\lambda$ is relatively small, the relative signal fluctuation gets stronger, which can lead to more noise events. Consequently, the analog circuit of an event pixel is designed to have a photoreceptor bias voltage and source-follower buffer which help stabilize the photoreceptor's reading and filter out fluctuations beyond a certain bandwidth~\cite{delbruck1995analog, gracca2023shining}. To account for this in our model, we approximate the filtering effect by adding a photoreceptor bias term, $b_{pr}$, to the photon count before the logarithmic operation. 
As a result, the triggering condition of a positive event becomes $\log(n + b_{pr}) - \log(n_0 + b_{pr}) > \epsilon$, which can be re-written as $n + b_{pr} - e^\epsilon \left( n_0 + b_{pr}\right) > 0$. We can similarly write the probability of triggering an event, $p_e$, as a function of the average photon count (proportional to the illuminance):
\begin{equation}
p_e\left( \lambda \right) = \frac{1}{2} - \frac{1}{2} \textit{erf} \left( \frac{\left(\lambda + b_{pr} \right) \left( e^\epsilon - 1\right) }{\sqrt{2 \lambda \left( 1 + e^{2\epsilon} \right)}} \right). \label{eq-forward}
\end{equation}
With the bias term, the model shows fewer noise events at lower illuminance levels (Fig.~\ref{fig-theory-curve}a). As average photon count increases, the number of noise events will increase up to a point and then decrease. While this model is not monotonic, we will show that we can still recover illuminance from noise event counts.

To validate our derivation, we acquired experimental measurements of noise events at different illuminance levels (Fig.~\ref{fig-theory-curve}b). The measured positive event rate matches to our derived noise model after parameter fitting. Note that our formulation is a simplified noise model for the event circuit, which characterizes the events triggered by photon noise. There are more controllable bias parameters in the actual analog circuit~\cite{delbruck1995analog,gracca2023shining} that are beyond our formulation.

\begin{figure}[h] 
\centering
\includegraphics[width=0.8\textwidth]{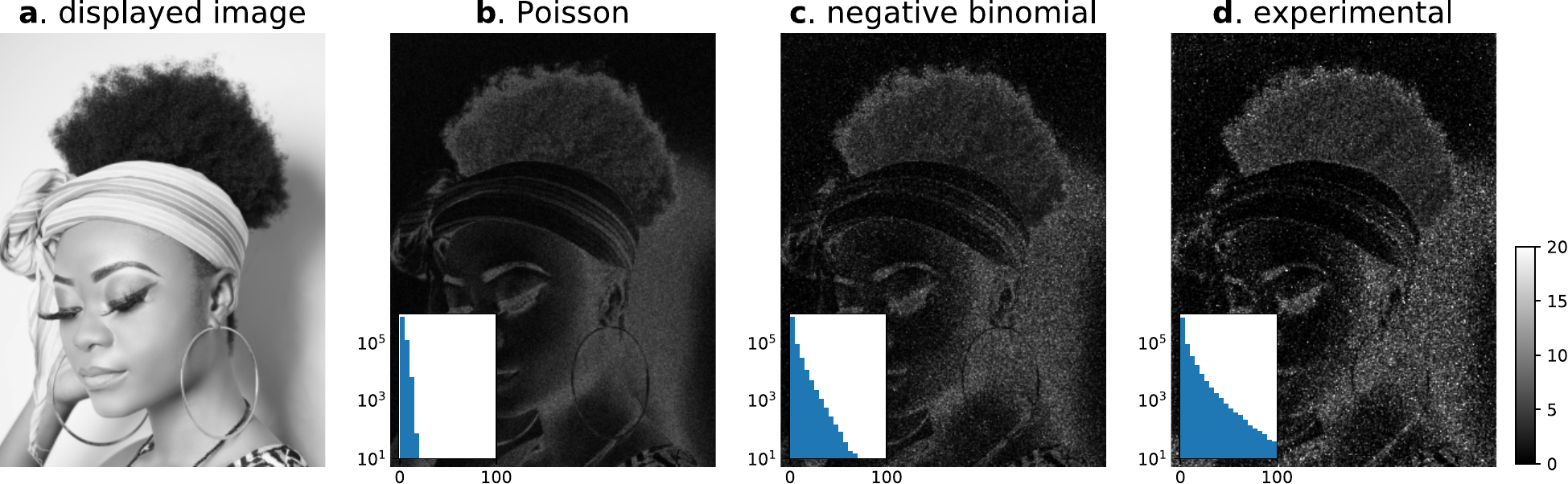}  
\caption{\textbf{a}. Displayed static scene. \textbf{b-c}. Synthetic noise event count sampled from the Poisson distribution and generalized negative binomial distribution, respectively. \textbf{d}. Noise event count from a experimental recording. The insets show the histogram for the noise event count. The frequency of the histogram (y-axis) is in log scale.} \label{fig-synthesis}
\end{figure}

\subsection{Sampling Noise Events}
\label{sec-noise_sampling}
Once we know the probability of observing a single noise event, we can simulate the number of noise events over a fixed time window given the intensity of a scene, $I$. Viewing each potential noise event in this window as a Bernoulli trial with probability $p_e$, the noise event count has a binomial distribution with probability of $p_e\left( I \right)$ and $N$ trials. With the refractory period (\ie, minimum time between events) much smaller than the inter-event interval (\ie, the time interval between two consecutive noise events at the same pixel), $N$ will be large, and the binomial distribution can be approximated by a Poisson distribution with parameter $N \cdot p_e\left( I \right)$.

However, we observe that the empirical noise event count (Fig.~\ref{fig-synthesis}d) has a higher variance than the Poisson distribution (Fig.~\ref{fig-synthesis}b), referred to as overdispersion. The overdispersion of the experimental data is likely caused by some inter-pixel variabilities, such as the variation of contrast threshold values~\cite{Lichtsteiner2008A11}. Thus, we instead sample the noise count from a generalized negative binomial distribution---which is often used in lieu of Poisson when there is overdispersion---with a mean of $N \cdot p_e\left( I \right)$ and an illuminance-dependent variance obtained empirically from the calibration data. The noise count sampled from the negative binomial distribution (Fig.~\ref{fig-synthesis}c) resembles the experimental count.

\section{Reconstructing the Scene}
Once we have a model for the mapping between scene intensity and noise event count at each pixel, we can develop an algorithm for recovering the static scene. We first estimate the true noise event count $N \cdot p_e\left( I \right)$ from the empirical noise event count that the camera measures, and then estimate the intensity $I$ by inverting Eq.~\eqref{eq-forward}. By doing this for every pixel, we can form an image for a static scene.

This inverse problem is non-trivial to solve for two reasons. First, the problem is ill-posed since Eq.~\eqref{eq-forward}  is one-to-many, as in Fig.~\ref{fig-theory-curve}b, for nonzero biases and thus using the noise count alone is insufficient to solve for exact intensity values. Second, even though the empirical event count is the maximum likelihood estimator of the true event count, the estimate from the empirical event counts in a finite time window will have some statistical error leading to further downstream error after inverting $p_e(I)$. 

We approach the first challenge by counting the positive and negative polarities of noise events separately. We found that event cameras often set different contrast threshold values to trigger positive and negative events, since the leakage current of the event circuit causes positive events to be triggered more easily~\cite{nozaki2017temperature}. Because of potentially asymmetric behavior of the two polarities, the correlation between light intensity and noise events is also different for each polarity. As shown in Fig.~S1, the corresponding light intensity becomes more uniquely defined when counting positive and negative events separately. 

The second issue --- inexact event count estimation --- can also be mitigated through the use of data priors in the inverse problem. This will make the resulting reconstruction robust to small errors in the event rate. Combining these two, we train a neural network to map the estimated event count directly to the corresponding intensity image. The network accepts inputs with two spatial channels for positive and negative event counts. To further reduce the variance, we perform pixel binning across blocks of $2\times2$ pixels.

\subsection{Application on Dynamic Scenes}
As in Fig.~\ref{fig-pipeline}f, the static scene reconstruction can also operate in parallel with the event-to-video reconstruction (E2VID) pipeline on recordings with both static and dynamic components, \eg, scenes with moving foreground and static background. We first separate signal events triggered by dynamic changes from other recorded events using an event denoising algorithm~\cite{feng2020event, ding2023mlb}, and apply an E2VID method to reconstruct the dynamic components. The remaining events can be considered noise events triggered mostly by photon fluctuation, which can be fed into the proposed Noise2Image model for static background reconstruction. For a variable-length recording, we aggregate noise events using a moving window with a fixed temporal width. Lastly, we stitch the dynamic and static scene reconstructions together using a binary motion mask identified from the signal events.

\section{Experiments}
\label{sec-exp}

The existing event camera datasets all have a significant amount of scene motion and/or camera motion. It is not possible to fully distinguish between events triggered by intensity changes and noise events~\cite{Baldwin_2020_CVPR}, and thus they are not suitable for our goal of static scene reconstruction. Hence, we collect our own training/validation dataset, termed noise events-to-image (NE2I). NE2I contains pairs of high-resolution intensity images and noise event recordings from both experimental acquisition and synthetic noise based on the model presented in Sec.~\ref{sec-noise_sampling}. 

We image a 24.5 inch LCD monitor displaying static images. In order to calibrate the noise model, we display 256 grayscale values on the monitor and capture both their event response and light intensity. The illuminance of each grayscale value is measured using a light meter placed next to the camera. Our event camera is a monochromatic Prophesee Metavision EVK3-HD. The default bias parameters were used for the camera, and no denoising was performed on the raw data. While the event camera has a high pixel count ($1280\times720$ pixels), it does not have an active pixel sensor (APS) that records frame-based intensity images. To register the event camera recording to the screen, we first imaged a standard checkerboard flashing on the monitor to induce events. The aggregated event count is used to establish a transformation matrix~\cite{opencv_library}, which is applied to any event recordings to align it spatially with the monitor.

The full NE2I dataset consists of an in-distribution set and an out-of-distribution test set. The in-distribution data contains 1004 high-resolution images of artistic human portraits from Unsplash, split into 754 images for training, 100 for validation, and 150 for testing. The out-of-distribution test set has 100 high-resolution images from the validation set of DIV2K image super-resolution dataset~\cite{agustsson2017ntire}, aiming to provide a variety of scenes much beyond the training data distribution. We intentionally chose a confined distribution for training data and a broad distribution for testing, so that the evaluation can test if the model learns local correlations instead of an image-level prior.

Given a sequence of noise events, we first aggregate them into a 2D-matrix of event count. To account for the positive and negative polarities, we separately aggregate them and store as two channels. We train a U-net to map an event count matrix into the corresponding intensity image. A modified U-net from~\cite{ho2020denoising} is used for enhanced performance. The network is trained using either experimental training data (described above) or synthetic event counts generated by the statistical noise model in Sec.~\ref{sec-noise_sampling}. For experimental data training, we augment the input noise count by choosing a random starting time for the 1-second window over a 10-second event recording. For synthetic data training, event counts are re-sampled each time on-the-fly. After training, the model takes around 100ms to recover a static scene on an NVIDIA RTX3090 GPU.

We use E2VID as a baseline method for the evaluation of static scene recovery even though it was not trained with such data. Rather than quantifying our performance gain, this comparison aims to test whether E2VID can identify the correlation between noise events and illuminance level. We compare to the pre-trained model implementation~\cite{ercan2023evreal} for the original E2VID~\cite{rebecq2019high}, E2VID+~\cite{stoffregen2020reducing}, FireNet~\cite{Scheerlinck20wacv}, FireNet+~\cite{stoffregen2020reducing}, ET-Net~\cite{weng2021event}, SPADE-E2VID~\cite{cadena2021spade}, SSL-E2VID~\cite{paredes2021back}, HyperE2VID~\cite{ercan2023hypere2vid}. A sequence of events within a 1-second window is fed into each method, and the last frame of predicted video (5 predicted frames in total) is used for evaluation.

\subsection{Metrics}
Evaluation of all methods is performed using experimentally-collected testing data, with the quality of recovered intensity images being calculated for the in-distribution and out-of-distribution testing datasets using three common quantitative metrics~\cite{rebecq2019high}: peak signal-to-noise ratio (PSNR), structured similarity (SSIM) and perceptual similarity (LPIPS~\cite{zhang2018perceptual}). LPIPS is computed using pre-trained AlexNet with image intensity values normalized to [-1, 1]. 

\subsection{Results}

\begin{table}[]
\centering
\begin{tabular}{ p{4cm} p{1cm}p{1cm}p{1cm} c p{1cm}p{1cm}p{1cm} }
 \hline
 \multirow{2}{*}{Methods} & \multicolumn{3}{c}{In-distribution test} && \multicolumn{3}{c}{Out-of-distribution test} \\
 \cline{2-4} \cline{6-8}
  & PSNR & SSIM & LPIPS & & PSNR & SSIM & LPIPS \\
 \hline
 Original E2VID~\cite{rebecq2019high}            & 7.76 & 0.043 & 1.131 && 10.1 & 0.055 & 1.032 \\
 E2VID+~\cite{stoffregen2020reducing}   & 8.72 & 0.148 & 0.872 && 9.26 & 0.108 & 0.768 \\
 FireNet~\cite{Scheerlinck20wacv}       & 9.35 & 0.089 & 1.047 && 10.5 & 0.098 & 0.970 \\
 FireNet+~\cite{stoffregen2020reducing} & 9.78 & 0.063 & 0.959 && 9.77 & 0.057 & 0.800 \\
 ET-Net~\cite{weng2021event}            & 9.13 & 0.166 & 0.805 && 10.1 & 0.131 & 0.730 \\
 SPADE-E2VID~\cite{cadena2021spade}     & 7.74 & 0.229 & 0.856 && 8.74 & 0.149 & 0.870 \\
 SSL-E2VID~\cite{paredes2021back}       & 8.81 & 0.133 & 0.764 && 8.95 & 0.123 & 0.720 \\
 HyperE2VID~\cite{ercan2023hypere2vid}  & 9.09 & 0.159 & 0.858 && 0.89 & 0.094 & 0.759 \\
 \hline
 Noise2Image, synthetic                    & 20.3 & 0.614 & 0.449 && 16.6 & 0.504 & 0.532 \\
 Noise2Image, experimental                   & 25.0 & 0.742 & 0.349 && 19.3 & 0.552 & 0.509 \\
 \hline
\end{tabular}
\caption{Quantitative results for static scene reconstruction with in-distribution and out-of-distribution testing data. Our Noise2Image models are trained with either synthetic or experimental noise data. Pre-trained events-to-video reconstruction (E2VID) methods are used as baselines. We report peak signal-to-noise ratio (PSNR), structured similarity (SSIM), and perceptual similarity (LPIPS~\cite{zhang2018perceptual}). We note that since E2VID models were trained by synthetic motion data generated from the MS-COCO dataset~\cite{lin2014microsoft}, both of our testing sets are considered out-of-distribution.}
\label{table-quant}
\end{table}

\begin{figure}[]
\centering
\includegraphics[width=0.7\textwidth]{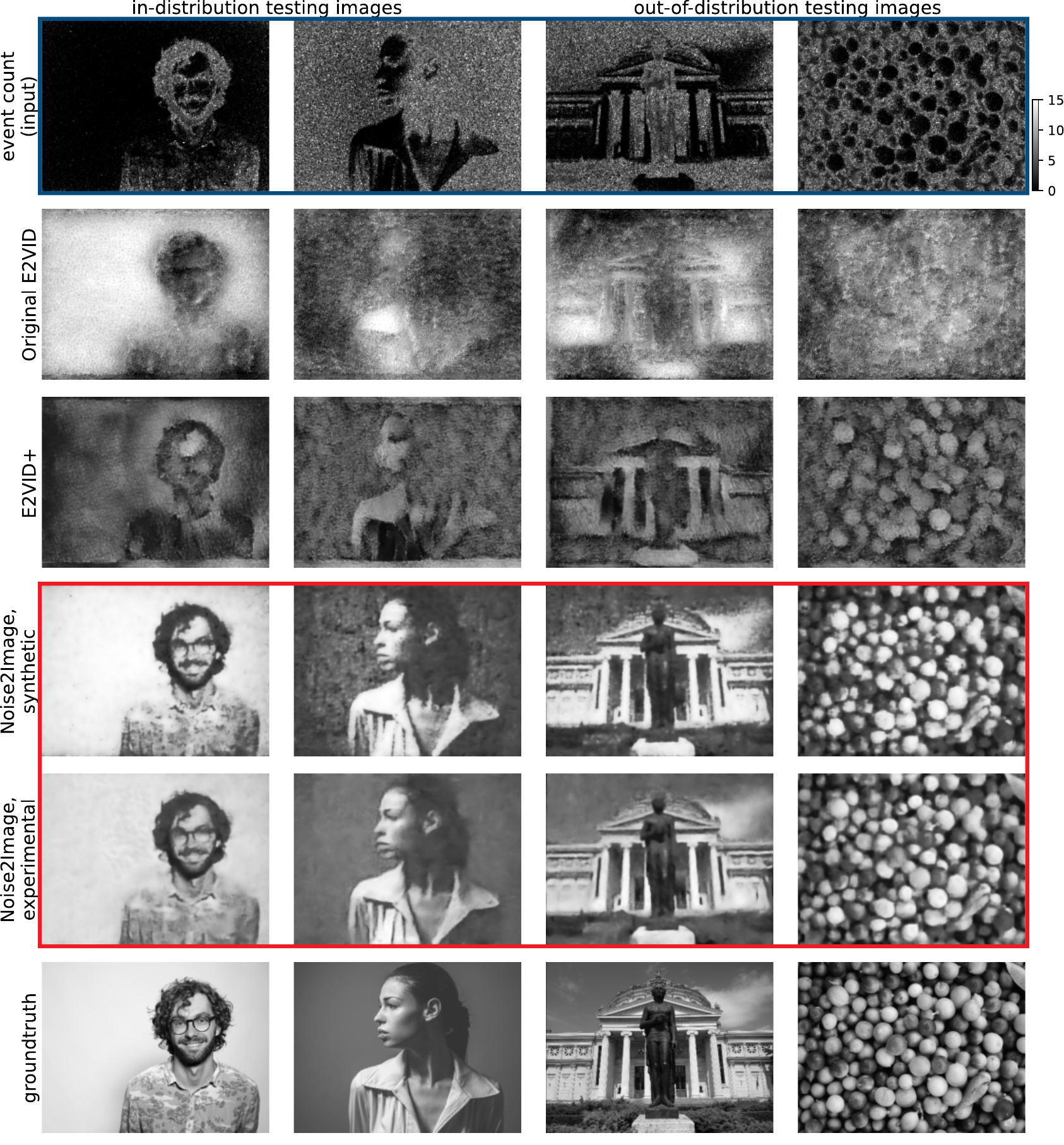}
\caption{Comparison between Noise2Image and baseline pre-trained event-to-video (E2VID) methods on noise event-to-intensity reconstruction. The first row shows the input event count (captured in experiment) aggregated over a 1-second window. Noise2Image is trained using either synthetic or experimental data as specified. Full comparison with all baseline methods can be found in Fig.~S2. } \label{fig-qualitative}
\end{figure}

For static-only scenes, the quantitative scores for Noise2Image as well as baseline methods using NE2I dataset are reported in Table~\ref{table-quant}. As shown in Fig.~\ref{fig-qualitative}, the Noise2Image model trained by either experimental or synthetic data recovers detailed intensity images from the count of noise events. Both Noise2Image models generalized well to out-of-distribution testing data, providing a good level of contrast and details, suggesting that Noise2Image learns the correlation between noise event and intensity. 
The Noise2Image model trained by experimental data out-performed the one trained by the synthetic data by 4.1dB of PSNR. As the synthetic data-trained model often recovered speckle-like patterns for the uniform image background, we speculate that this performance margin is caused by the variability of contrast thresholds across pixels, which is beyond our synthesis model but potentially captured by the experimental data training. 
Baseline E2VID methods did not perform well on static scene reconstruction for two reasons: they were only trained by scenes with various degrees of motion (not static scenes) and the training data of E2VID was generated without an intensity-dependent noise model~\cite{rebecq2018esim}.

\begin{figure}[]
\centering
\includegraphics[width=0.7\textwidth]{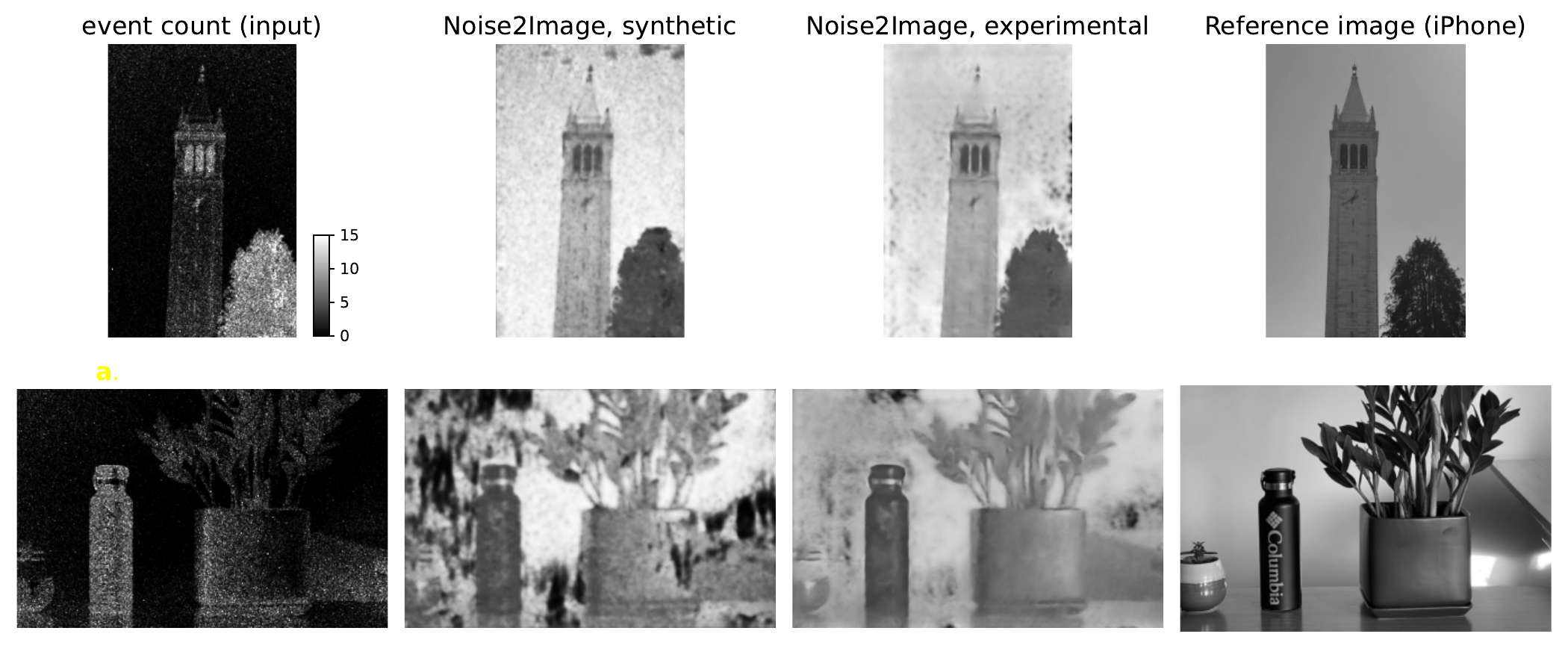}  
\caption{Real-world examples of Noise2Image taken outside of the laboratory setting. The Noise2Image model trained by synthetic data results in background artifacts on the in-door scene (second row, second column), presumably caused by the one-to-many relationship of Eq.~3. In contrast, the Noise2Image trained by experimental data predicts the background correctly, hinting that there exist spatial structures in the experimental noise sensitivity, beyond our derived spatially independent noise event synthesis. The reference images were taken by an iPhone 12 plus back camera. } \label{fig-outdoor_results}
\end{figure}

In addition to testing data, we tested Noise2Image on real-world scenes, both indoor and outdoor. The Noise2Image model trained by images displayed on a monitor generalizes well for scenes outside of the laboratory setting, as shown in Fig.~\ref{fig-outdoor_results}. We have implemented a real-time demo running on a laptop computer as well.

Table~S1 shows the effect of varying the aggregation window time for training and testing the Noise2Image method. As the aggregation duration shortens, fewer noise events will be triggered, and the estimation of the true event count becomes less accurate. Noise2Image reconstruction works even with short aggregation duration (0.1 seconds), although longer integration will result in better reconstruction quality. We also tested using only events from a single polarity, which resulted in a slight reduction in performance as shown in Table~\ref{table-polarity}.

\begin{table}
\centering
\begin{tabular}{ p{3cm}p{1cm}p{1cm}p{1cm}c p{1cm}p{1cm}p{1cm} }
\hline
\multirow{2}{*}{Event polarity} & \multicolumn{3}{c}{In-distribution test} & & \multicolumn{3}{c}{Out-of-distribution test} \\
\cline{2-4} \cline{6-8}
                    & PSNR & SSIM & LPIPS && PSNR & SSIM & LPIPS \\
\hline
Positive \& negative               & 25.0 & 0.742 & 0.349 && 19.3 & 0.552 & 0.509 \\
Positive only              & 24.1 & 0.686 & 0.378 && 18.7 & 0.501 & 0.526 \\
Negative only              & 24.2 & 0.696 & 0.378 && 18.8 & 0.513 & 0.531 \\
\hline
\end{tabular}
\caption{Ablation study of event polarities. We train and test Noise2Image method using both positive and negative events, positive events only, and negative events only. }
\label{table-polarity}
\end{table}

\subsection{Reconstructing scenes with dynamic components}
\begin{figure}[]
\centering
\includegraphics[width=0.8\textwidth]{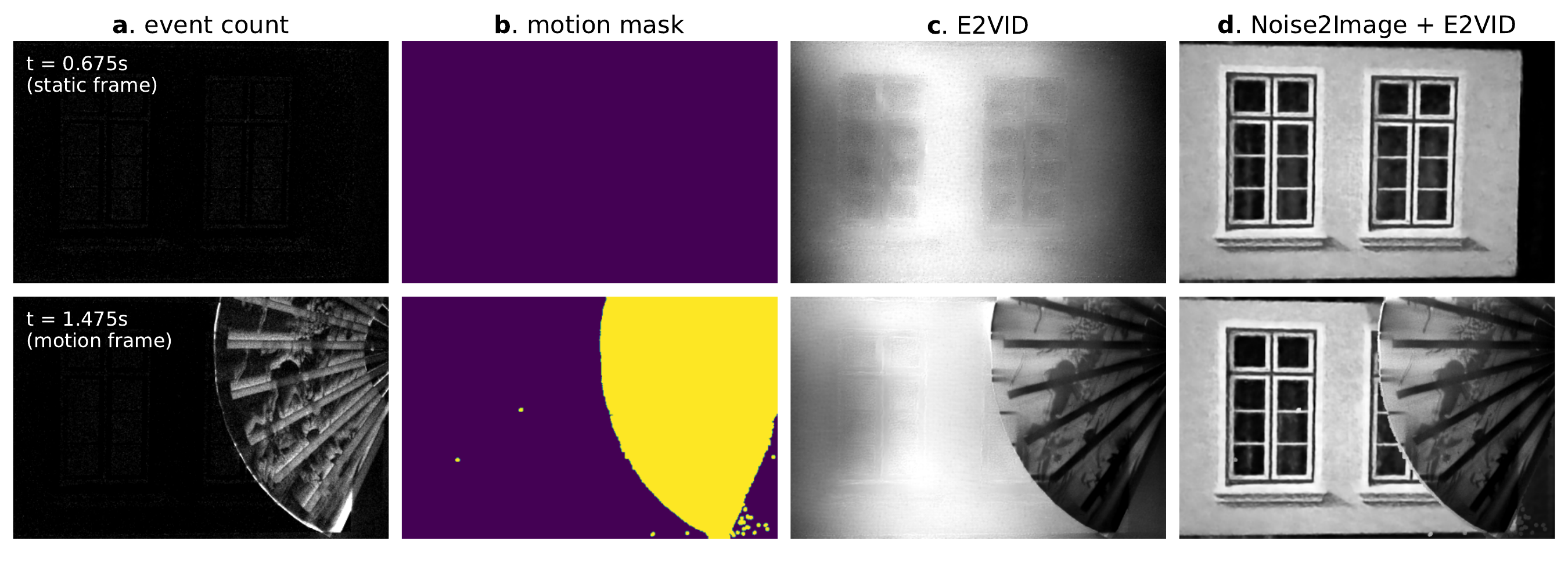}
\caption{Dynamic scene reconstruction of a moving fan in front of a static scene (see Visualization 1). \textbf{a}. Aggregated event count. The first row is at a timepoint with no motion and only faint noise, and the second row has motion. \textbf{b}. Motion mask obtained from thresholding the signal events as determined by the event denoiser (YNoise denoiser from \cite{ynoise} implemented in \cite{ding2023mlb}). \textbf{c}. E2VID reconstruction using pre-trained model from~\cite{rebecq2019high}. \textbf{d}. The dynamic foreground reconstructed by E2VID and the static background reconstructed by Noise2Image are stitched together. } \label{fig-motion}
\end{figure}

Finally, we demonstrated that Noise2Image is complementary to E2VID reconstruction when the scene has both static and dynamic components. We imaged a fan rapidly moving in front of a static scene, and fed signal events into a pre-trained E2VID model~\cite{rebecq2019high}. While the E2VID method performed well on the moving object, it could not recover the background static scene (Fig.~\ref{fig-motion}c). To incorporate Noise2Image, we first identified a motion mask at each timepoint (Fig.~\ref{fig-motion}b) by thresholding signal events. We then aggregated noise events and fed the event count at pixels without motion to the Noise2Image model. Stitching together the moving foreground from E2VID and the static background from Noise2Image, we obtained the final high-quality reconstruction as in Fig.~\ref{fig-motion}d and Visualization 1.

\section{Discussion}
Our experiments were done using a single event camera, and there might be differences in the noise characteristics between different event circuit designs. 
However, our finding in Sec.~\ref{sec-theory} is based on the photon arrival process, which is generalizable to other event camera hardware. This correlation between noise events and illuminance level was also independently reported in previous studies using different event cameras~\cite{graca2021unraveling, hu2021v2e, ding2023mlb, gracca2023shining}.

Denoising and camera bias parameters~\cite{mcreynolds2023demystifying} can also affect this illuminance dependency. Our data was acquired without denoising and using bias parameters default to our event camera. We also tested different high-pass filtering bias values which affect the correlation between noise event rate and illuminance (Fig.~S3). 
For this reason, the trained Noise2Image model is bound to the camera model and bias parameters that were used in the training data collection. 
We hope that our modeling of noise events can be improved upon by incorporating the analog nature of event sampling and modeling additional event camera bias parameters, \eg, using non-parametric models to account for unknown camera behavior. 

In our evaluation, Noise2Image trained by experimental data outperforms the one trained by synthetic data, which suggests discrepancy between the captured noise distribution and our theoretical noise model. From our observation, there exist some spatial structures in the experimental noise event count which is beyond our theoretical noise model. We believe this is caused by the variability of contrast threshold, a known issue related to the manufacturing of event sensor~\cite{Lichtsteiner2008A11, wang2020eventbias}.

The dynamic range of Noise2Image is limited to the moderate-brightness regime, where noise events are abundant. Although our theoretical noise model still holds in the high-brightness regime, there are fewer noise events, providing little signal for static scene recovery. 
Our empirical study is further constrained by the dynamic range of the monitor we used, suggesting that a high-dynamic-range, high-brightness monitor or projector could enhance the dynamic range of the experimental training data. 
We hypothesize that our Noise2Image approach may still work for brighter scenes with the presence of leakage current-induced noise events~\cite{nozaki2017temperature}, and will leave this for future research. For low-brightness scenes, it is worth further investigating the feasibility of the Noise2Image approach under extremely low-signal conditions, such as in fluorescence microscopy.

Another future direction is to incorporate the proposed noise event modeling into existing event camera simulators. This will help synthesize data with realistic noise events, which can be used for training event-to-video recovery (E2VID) models. Overall, the model similarities between Noise2Image and E2VID suggest that in the future, Noise2Image can be incorporated into the E2VID pipeline via more realistic training data, complementing each other's goals.

In summary, we demonstrated imaging of static intensity using an event camera with no additional hardware. This is made possible by the fact that photon-noise triggers events which are correlated with illuminance. We developed a statistical model of the noise event generation, and leveraged it to develop a strategy called Noise2Image, which maps noise events to an intensity image via a neural network that incorporates local priors. We demonstrate that our approach works well quantitatively and qualitatively on experimental event recordings, including ones taken in the wild. We also show that our method is compatible with scenes that have object dynamics. Our hope is that Noise2Image reduces the need for event cameras to have additional hardware for measuring static scenes.

\section*{Disclosure}
Authors declare no conflict of interest.

\section*{Code and Data Availability}
Code and data presented in this paper are available: \href{https://github.com/rmcao/noise2image}{https://github.com/rmcao/noise2image}.

\section*{Acknowledgments}
We thank Joshua Follansbee for the feedback on our derivations. This material is based upon work supported by the National Science Foundation Graduate Research Fellowship Program under Grant No. DGE-1752814. Any opinions, findings, and conclusions or recommendations expressed in this material are those of the author(s) and do not necessarily reflect the views of the National Science Foundation. Research was supported by the National Eye Institute of the National Institute of Health under award number R00EY032179. Ruiming Cao was supported in part by Siebel Scholarship. This work was supported by Weill Neurohub Investigators Program. This work has been made possible in part by CZI grant DAF2021-225666 and grant DOI \href{https://doi.org/10.37921/192752jrgbnh}{10.37921/192752jrgbnh} from the Chan Zuckerberg Initiative DAF, an advised fund of Silicon Valley Community Foundation (funder DOI 10.13039/100014989). Laura Waller is a Chan Zuckerberg Biohub SF Investigator.

\bibliography{main}

\begin{thebibliography}{10}
\newcommand{\enquote}[1]{``#1''}

\bibitem{mahowald1994silicon}
M.~Mahowald and M.~Mahowald, \enquote{The silicon retina,} {\protect\JournalTitle{An Analog VLSI System for Stereoscopic Vision}} pp. 4--65 (1994).

\bibitem{gallego2020event}
G.~Gallego, T.~Delbr{\"u}ck, G.~Orchard, \emph{et~al.}, \enquote{Event-based vision: A survey,} {\protect\JournalTitle{IEEE transactions on pattern analysis and machine intelligence}} \textbf{44}, 154--180 (2020).

\bibitem{angelopoulos2020event}
A.~N. Angelopoulos, J.~N. Martel, A.~P. Kohli, \emph{et~al.}, \enquote{Event based, near eye gaze tracking beyond 10,000 hz,} {\protect\JournalTitle{IEEE Transactionson Visualizations and Graphics}}  (2021).

\bibitem{brandli2014real}
C.~Brandli, L.~Muller, and T.~Delbruck, \enquote{Real-time, high-speed video decompression using a frame-and event-based davis sensor,} in \emph{2014 IEEE International Symposium on Circuits and Systems (ISCAS),}  (IEEE, 2014), pp. 686--689.

\bibitem{berner2013240}
R.~Berner, C.~Brandli, M.~Yang, \emph{et~al.}, \enquote{A 240$\times$ 180 10mw 12us latency sparse-output vision sensor for mobile applications,} in \emph{2013 Symposium on VLSI Circuits,}  (IEEE, 2013), pp. C186--C187.

\bibitem{hidalgo2022event}
J.~Hidalgo-Carri{\'o}, G.~Gallego, and D.~Scaramuzza, \enquote{Event-aided direct sparse odometry,} in \emph{Proceedings of the IEEE/CVF Conference on Computer Vision and Pattern Recognition,}  (2022), pp. 5781--5790.

\bibitem{zou2021learning}
Y.~Zou, Y.~Zheng, T.~Takatani, and Y.~Fu, \enquote{Learning to reconstruct high speed and high dynamic range videos from events,} in \emph{Proceedings of the IEEE/CVF Conference on Computer Vision and Pattern Recognition,}  (2021), pp. 2024--2033.

\bibitem{zhu2021neuspike}
L.~Zhu, J.~Li, X.~Wang, \emph{et~al.}, \enquote{Neuspike-net: High speed video reconstruction via bio-inspired neuromorphic cameras,} in \emph{Proceedings of the IEEE/CVF International Conference on Computer Vision,}  (2021), pp. 2400--2409.

\bibitem{wang2021asynchronous}
Z.~Wang, Y.~Ng, C.~Scheerlinck, and R.~Mahony, \enquote{An asynchronous kalman filter for hybrid event cameras,} in \emph{Proceedings of the IEEE/CVF International Conference on Computer Vision,}  (2021), pp. 448--457.

\bibitem{he2024microsaccade}
B.~He, Z.~Wang, Y.~Zhou, \emph{et~al.}, \enquote{Microsaccade-inspired event camera for robotics,} {\protect\JournalTitle{Science Robotics}} \textbf{9}, eadj8124 (2024).

\bibitem{guo2022low}
S.~Guo and T.~Delbruck, \enquote{Low cost and latency event camera background activity denoising,} {\protect\JournalTitle{IEEE Transactions on Pattern Analysis and Machine Intelligence}} \textbf{45}, 785--795 (2022).

\bibitem{gracca2023shining}
R.~Gra{\c{c}}a, B.~McReynolds, and T.~Delbruck, \enquote{Shining light on the dvs pixel: A tutorial and discussion about biasing and optimization,} in \emph{Proceedings of the IEEE/CVF Conference on Computer Vision and Pattern Recognition,}  (2023), pp. 4044--4052.

\bibitem{graca2021unraveling}
R.~Graca and T.~Delbruck, \enquote{Unraveling the paradox of intensity-dependent dvs pixel noise,} {\protect\JournalTitle{arXiv preprint arXiv:2109.08640}}  (2021).

\bibitem{nozaki2017temperature}
Y.~Nozaki and T.~Delbruck, \enquote{Temperature and parasitic photocurrent effects in dynamic vision sensors,} {\protect\JournalTitle{IEEE Transactions on Electron Devices}} \textbf{64}, 3239--3245 (2017).

\bibitem{scheerlinck2018continuous}
C.~Scheerlinck, N.~Barnes, and R.~Mahony, \enquote{Continuous-time intensity estimation using event cameras,} in \emph{Asian Conference on Computer Vision,}  (Springer, 2018), pp. 308--324.

\bibitem{jiang2020learning}
Z.~Jiang, Y.~Zhang, D.~Zou, \emph{et~al.}, \enquote{Learning event-based motion deblurring,} in \emph{Proceedings of the IEEE/CVF Conference on Computer Vision and Pattern Recognition,}  (2020), pp. 3320--3329.

\bibitem{pan2019bringing}
L.~Pan, C.~Scheerlinck, X.~Yu, \emph{et~al.}, \enquote{Bringing a blurry frame alive at high frame-rate with an event camera,} in \emph{Proceedings of the IEEE/CVF Conference on Computer Vision and Pattern Recognition,}  (2019), pp. 6820--6829.

\bibitem{wang2020event}
B.~Wang, J.~He, L.~Yu, \emph{et~al.}, \enquote{Event enhanced high-quality image recovery,} in \emph{Computer Vision--ECCV 2020: 16th European Conference, Glasgow, UK, August 23--28, 2020, Proceedings, Part XIII 16,}  (Springer, 2020), pp. 155--171.

\bibitem{wang2019event}
Z.~W. Wang, W.~Jiang, K.~He, \emph{et~al.}, \enquote{Event-driven video frame synthesis,} in \emph{Proceedings of the IEEE/CVF International Conference on Computer Vision Workshops,}  (2019), pp. 0--0.

\bibitem{tulyakov2021time}
S.~Tulyakov, D.~Gehrig, S.~Georgoulis, \emph{et~al.}, \enquote{Time lens: Event-based video frame interpolation,} in \emph{Proceedings of the IEEE/CVF conference on computer vision and pattern recognition,}  (2021), pp. 16155--16164.

\bibitem{tulyakov2022time}
S.~Tulyakov, A.~Bochicchio, D.~Gehrig, \emph{et~al.}, \enquote{Time lens++: Event-based frame interpolation with parametric non-linear flow and multi-scale fusion,} in \emph{Proceedings of the IEEE/CVF Conference on Computer Vision and Pattern Recognition,}  (2022), pp. 17755--17764.

\bibitem{zhang2022unifying}
X.~Zhang and L.~Yu, \enquote{Unifying motion deblurring and frame interpolation with events,} in \emph{Proceedings of the IEEE/CVF Conference on Computer Vision and Pattern Recognition,}  (2022), pp. 17765--17774.

\bibitem{wang2023event}
Z.~Wang, F.~Hamann, K.~Chaney, \emph{et~al.}, \enquote{Event-based continuous color video decompression from single frames,} {\protect\JournalTitle{arXiv preprint arXiv:2312.00113}}  (2023).

\bibitem{bardow2016simultaneous}
P.~Bardow, A.~J. Davison, and S.~Leutenegger, \enquote{Simultaneous optical flow and intensity estimation from an event camera,} in \emph{Proceedings of the IEEE conference on computer vision and pattern recognition,}  (2016), pp. 884--892.

\bibitem{rebecq2019high}
H.~Rebecq, R.~Ranftl, V.~Koltun, and D.~Scaramuzza, \enquote{High speed and high dynamic range video with an event camera,} {\protect\JournalTitle{IEEE transactions on pattern analysis and machine intelligence}} \textbf{43}, 1964--1980 (2019).

\bibitem{stoffregen2020reducing}
T.~Stoffregen, C.~Scheerlinck, D.~Scaramuzza, \emph{et~al.}, \enquote{Reducing the sim-to-real gap for event cameras,} in \emph{Computer Vision--ECCV 2020: 16th European Conference, Glasgow, UK, August 23--28, 2020, Proceedings, Part XXVII 16,}  (Springer, 2020), pp. 534--549.

\bibitem{Scheerlinck20wacv}
C.~Scheerlinck, H.~Rebecq, D.~Gehrig, \emph{et~al.}, \enquote{Fast image reconstruction with an event camera,} in \emph{{IEEE} Winter Conf. Appl. Comput. Vis. {(WACV)},}  (2020), pp. 156--163.

\bibitem{weng2021event}
W.~Weng, Y.~Zhang, and Z.~Xiong, \enquote{Event-based video reconstruction using transformer,} in \emph{Proceedings of the IEEE/CVF International Conference on Computer Vision,}  (2021), pp. 2563--2572.

\bibitem{cadena2021spade}
P.~R.~G. Cadena, Y.~Qian, C.~Wang, and M.~Yang, \enquote{Spade-e2vid: Spatially-adaptive denormalization for event-based video reconstruction,} {\protect\JournalTitle{IEEE Transactions on Image Processing}} \textbf{30}, 2488--2500 (2021).

\bibitem{paredes2021back}
F.~Paredes-Vall{\'e}s and G.~C. de~Croon, \enquote{Back to event basics: Self-supervised learning of image reconstruction for event cameras via photometric constancy,} in \emph{Proceedings of the IEEE/CVF Conference on Computer Vision and Pattern Recognition,}  (2021), pp. 3446--3455.

\bibitem{ercan2023hypere2vid}
B.~Ercan, O.~Eker, C.~Saglam, \emph{et~al.}, \enquote{Hypere2vid: Improving event-based video reconstruction via hypernetworks,} {\protect\JournalTitle{arXiv preprint arXiv:2305.06382}}  (2023).

\bibitem{rebecq2018esim}
H.~Rebecq, D.~Gehrig, and D.~Scaramuzza, \enquote{Esim: an open event camera simulator,} in \emph{Conference on robot learning,}  (PMLR, 2018), pp. 969--982.

\bibitem{delbruck1995analog}
T.~Delbruck and C.~Mead, \enquote{Analog vlsi phototransduction by continuous-time, adaptive, logarithmic photoreceptor circuits,} {\protect\JournalTitle{Technical report}}  (1995).

\bibitem{Lichtsteiner2008A11}
P.~Lichtsteiner, C.~Posch, and T.~Delbr{\"u}ck, \enquote{A 128x128 120 db 15mus latency asynchronous temporal contrast vision sensor,} {\protect\JournalTitle{IEEE Journal of Solid-State Circuits}} \textbf{43}, 566--576 (2008).

\bibitem{ding2023mlb}
S.~Ding, J.~Chen, Y.~Wang, \emph{et~al.}, \enquote{E-mlb: Multilevel benchmark for event-based camera denoising,} {\protect\JournalTitle{IEEE Transactions on Multimedia}}  (2023).

\bibitem{hu2021v2e}
Y.~Hu, S.-C. Liu, and T.~Delbruck, \enquote{v2e: From video frames to realistic dvs events,} in \emph{Proceedings of the IEEE/CVF Conference on Computer Vision and Pattern Recognition,}  (2021), pp. 1312--1321.

\bibitem{delbruck2008frame}
T.~Delbruck \emph{et~al.}, \enquote{Frame-free dynamic digital vision,} in \emph{Proceedings of Intl. Symp. on Secure-Life Electronics, Advanced Electronics for Quality Life and Society,}  vol.~1 (Citeseer, 2008), pp. 21--26.

\bibitem{czech2016evaluating}
D.~Czech and G.~Orchard, \enquote{Evaluating noise filtering for event-based asynchronous change detection image sensors,} in \emph{2016 6th IEEE International Conference on Biomedical Robotics and Biomechatronics (BioRob),}  (IEEE, 2016), pp. 19--24.

\bibitem{padala2018noise}
V.~Padala, A.~Basu, and G.~Orchard, \enquote{A noise filtering algorithm for event-based asynchronous change detection image sensors on truenorth and its implementation on truenorth,} {\protect\JournalTitle{Frontiers in neuroscience}} \textbf{12}, 118 (2018).

\bibitem{benosman2013event}
R.~Benosman, C.~Clercq, X.~Lagorce, \emph{et~al.}, \enquote{Event-based visual flow,} {\protect\JournalTitle{IEEE transactions on neural networks and learning systems}} \textbf{25}, 407--417 (2013).

\bibitem{liu2015design}
H.~Liu, C.~Brandli, C.~Li, \emph{et~al.}, \enquote{Design of a spatiotemporal correlation filter for event-based sensors,} in \emph{2015 IEEE International Symposium on Circuits and Systems (ISCAS),}  (IEEE, 2015), pp. 722--725.

\bibitem{barrios2018less}
J.~Barrios-Avil{\'e}s, A.~Rosado-Mu{\~n}oz, L.~D. Medus, \emph{et~al.}, \enquote{Less data same information for event-based sensors: A bioinspired filtering and data reduction algorithm,} {\protect\JournalTitle{Sensors}} \textbf{18}, 4122 (2018).

\bibitem{Baldwin_2020_CVPR}
R.~W. Baldwin, M.~Almatrafi, V.~Asari, and K.~Hirakawa, \enquote{Event probability mask (epm) and event denoising convolutional neural network (edncnn) for neuromorphic cameras,} in \emph{Proceedings of the IEEE/CVF Conference on Computer Vision and Pattern Recognition (CVPR),}  (2020).

\bibitem{feng2020event}
Y.~Feng, H.~Lv, H.~Liu, \emph{et~al.}, \enquote{Event density based denoising method for dynamic vision sensor,} {\protect\JournalTitle{Applied Sciences}}  (2020).

\bibitem{krizhevsky2012imagenet}
A.~Krizhevsky, I.~Sutskever, and G.~E. Hinton, \enquote{Imagenet classification with deep convolutional neural networks,} {\protect\JournalTitle{Advances in neural information processing systems}} \textbf{25} (2012).

\bibitem{dosovitskiy2020image}
A.~Dosovitskiy, L.~Beyer, A.~Kolesnikov, \emph{et~al.}, \enquote{An image is worth 16x16 words: Transformers for image recognition at scale,} {\protect\JournalTitle{arXiv preprint arXiv:2010.11929}}  (2020).

\bibitem{serrano2015poker}
T.~Serrano-Gotarredona and B.~Linares-Barranco, \enquote{Poker-dvs and mnist-dvs. their history, how they were made, and other details,} {\protect\JournalTitle{Frontiers in neuroscience}} \textbf{9}, 481 (2015).

\bibitem{orchard2015converting}
G.~Orchard, A.~Jayawant, G.~K. Cohen, and N.~Thakor, \enquote{Converting static image datasets to spiking neuromorphic datasets using saccades,} {\protect\JournalTitle{Frontiers in neuroscience}} \textbf{9}, 437 (2015).

\bibitem{li2017cifar10}
H.~Li, H.~Liu, X.~Ji, \emph{et~al.}, \enquote{Cifar10-dvs: an event-stream dataset for object classification,} {\protect\JournalTitle{Frontiers in neuroscience}} \textbf{11}, 309 (2017).

\bibitem{kim2021n}
J.~Kim, J.~Bae, G.~Park, \emph{et~al.}, \enquote{N-imagenet: Towards robust, fine-grained object recognition with event cameras,} in \emph{Proceedings of the IEEE/CVF international conference on computer vision,}  (2021), pp. 2146--2156.

\bibitem{gehrig2020video}
D.~Gehrig, M.~Gehrig, J.~Hidalgo-Carri{\'o}, and D.~Scaramuzza, \enquote{Video to events: Recycling video datasets for event cameras,} in \emph{Proceedings of the IEEE/CVF Conference on Computer Vision and Pattern Recognition,}  (2020), pp. 3586--3595.

\bibitem{zhu2021eventgan}
A.~Z. Zhu, Z.~Wang, K.~Khant, and K.~Daniilidis, \enquote{Eventgan: Leveraging large scale image datasets for event cameras,} in \emph{2021 IEEE International Conference on Computational Photography (ICCP),}  (IEEE, 2021), pp. 1--11.

\bibitem{opencv_library}
G.~Bradski, \enquote{{The OpenCV Library},} {\protect\JournalTitle{Dr. Dobb's Journal of Software Tools}}  (2000).

\bibitem{agustsson2017ntire}
E.~Agustsson and R.~Timofte, \enquote{Ntire 2017 challenge on single image super-resolution: Dataset and study,} in \emph{Proceedings of the IEEE conference on computer vision and pattern recognition workshops,}  (2017), pp. 126--135.

\bibitem{ho2020denoising}
J.~Ho, A.~Jain, and P.~Abbeel, \enquote{Denoising diffusion probabilistic models,} {\protect\JournalTitle{Advances in neural information processing systems}} \textbf{33}, 6840--6851 (2020).

\bibitem{ercan2023evreal}
B.~Ercan, O.~Eker, A.~Erdem, and E.~Erdem, \enquote{Evreal: Towards a comprehensive benchmark and analysis suite for event-based video reconstruction,} in \emph{Proceedings of the IEEE/CVF Conference on Computer Vision and Pattern Recognition,}  (2023), pp. 3942--3951.

\bibitem{zhang2018perceptual}
R.~Zhang, P.~Isola, A.~A. Efros, \emph{et~al.}, \enquote{The unreasonable effectiveness of deep features as a perceptual metric,} in \emph{CVPR,}  (2018).

\bibitem{lin2014microsoft}
T.-Y. Lin, M.~Maire, S.~Belongie, \emph{et~al.}, \enquote{Microsoft coco: Common objects in context,} in \emph{Computer Vision--ECCV 2014: 13th European Conference, Zurich, Switzerland, September 6-12, 2014, Proceedings, Part V 13,}  (Springer, 2014), pp. 740--755.

\bibitem{ynoise}
Y.~Feng, H.~Lv, H.~Liu, \emph{et~al.}, \enquote{Event density based denoising method for dynamic vision sensor,} {\protect\JournalTitle{Applied Sciences}} \textbf{10} (2020).

\bibitem{mcreynolds2023demystifying}
B.~McReynolds, R.~Graca, R.~Oliver, \emph{et~al.}, \enquote{Demystifying event-based sensor biasing to optimize signal to noise for space domain awareness,} in \emph{Advanced Maui Optical and Space Surveillance Technologies Conference (AMOS),}  (University of Zurich, 2023).

\bibitem{wang2020eventbias}
Z.~Wang, Y.~Ng, P.~van Goor, and R.~Mahony, \enquote{Event camera calibration of per-pixel biased contrast threshold,} {\protect\JournalTitle{arXiv preprint arXiv:2012.09378}}  (2020).

\end{thebibliography}

\clearpage

\renewcommand{\figurename}{Supplementary Fig.}
\setcounter{figure}{0}

\begin{figure}[b]
\centering
\includegraphics[width=0.5\textwidth]{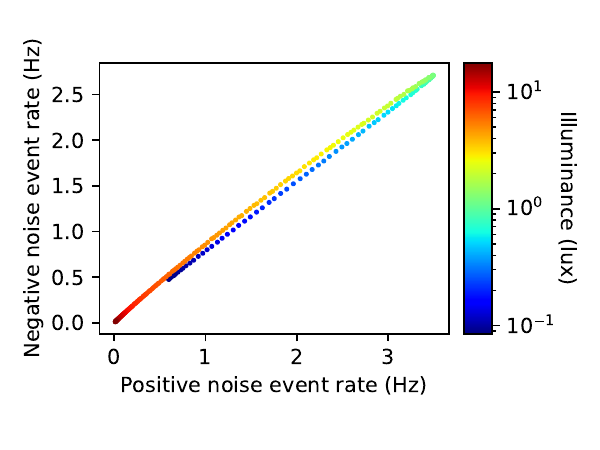}  
\caption{Noise event counts on positive events versus negative events. The color indicates the corresponding illuminance level. } \label{fig-pos_neg_count}
\end{figure}

\begin{figure}[]
\centering
\includegraphics[width=0.7\textwidth]{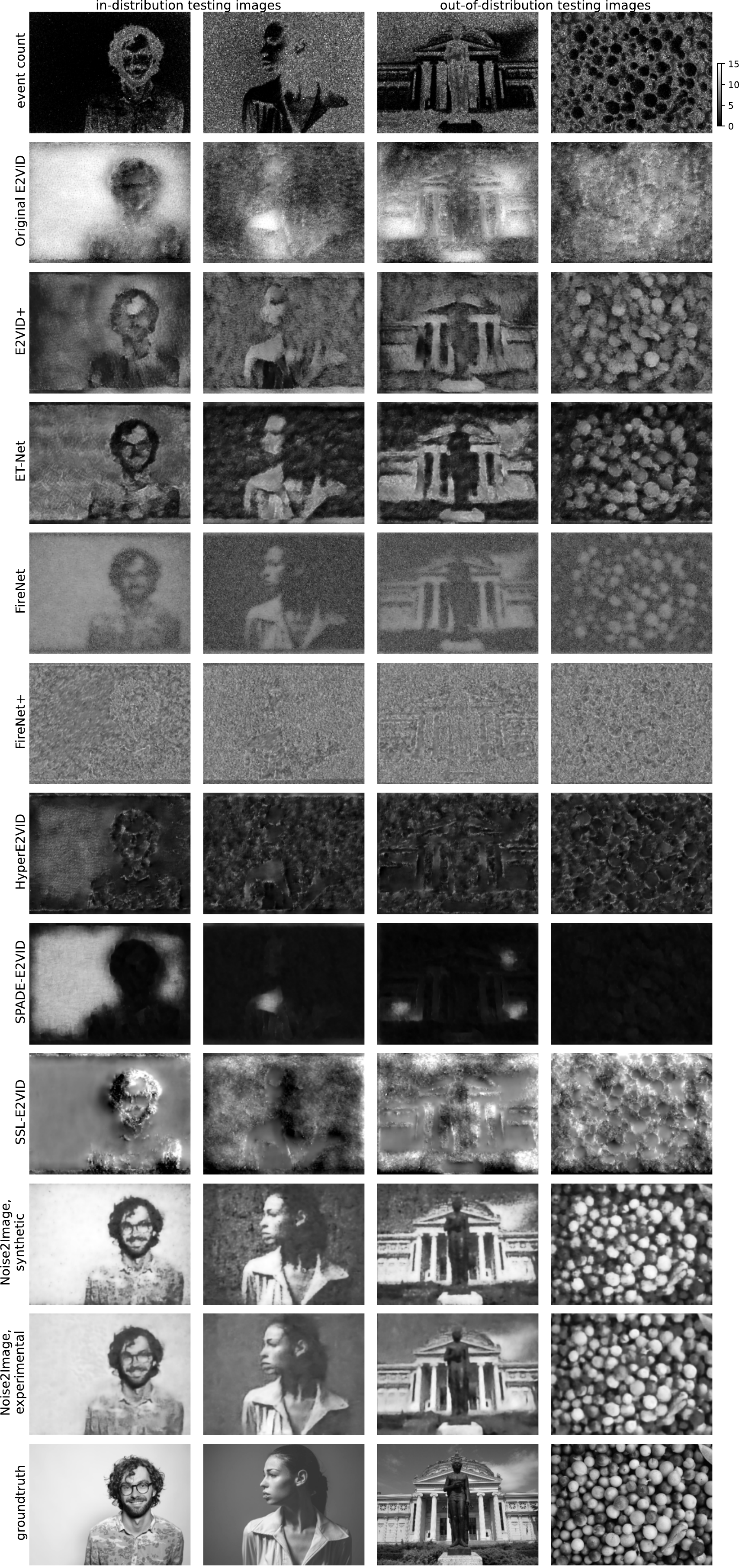}  
\caption{Additional comparison between Noise2Image and all baseline event-to-video (E2VID) methods on noise event-to-intensity reconstruction. The first row shows the input event count (captured in experiment) aggregated over a 1-second window.} \label{fig-qual_all_baselines}
\end{figure}

\begin{figure}[]
\centering
\includegraphics[width=0.5\textwidth]{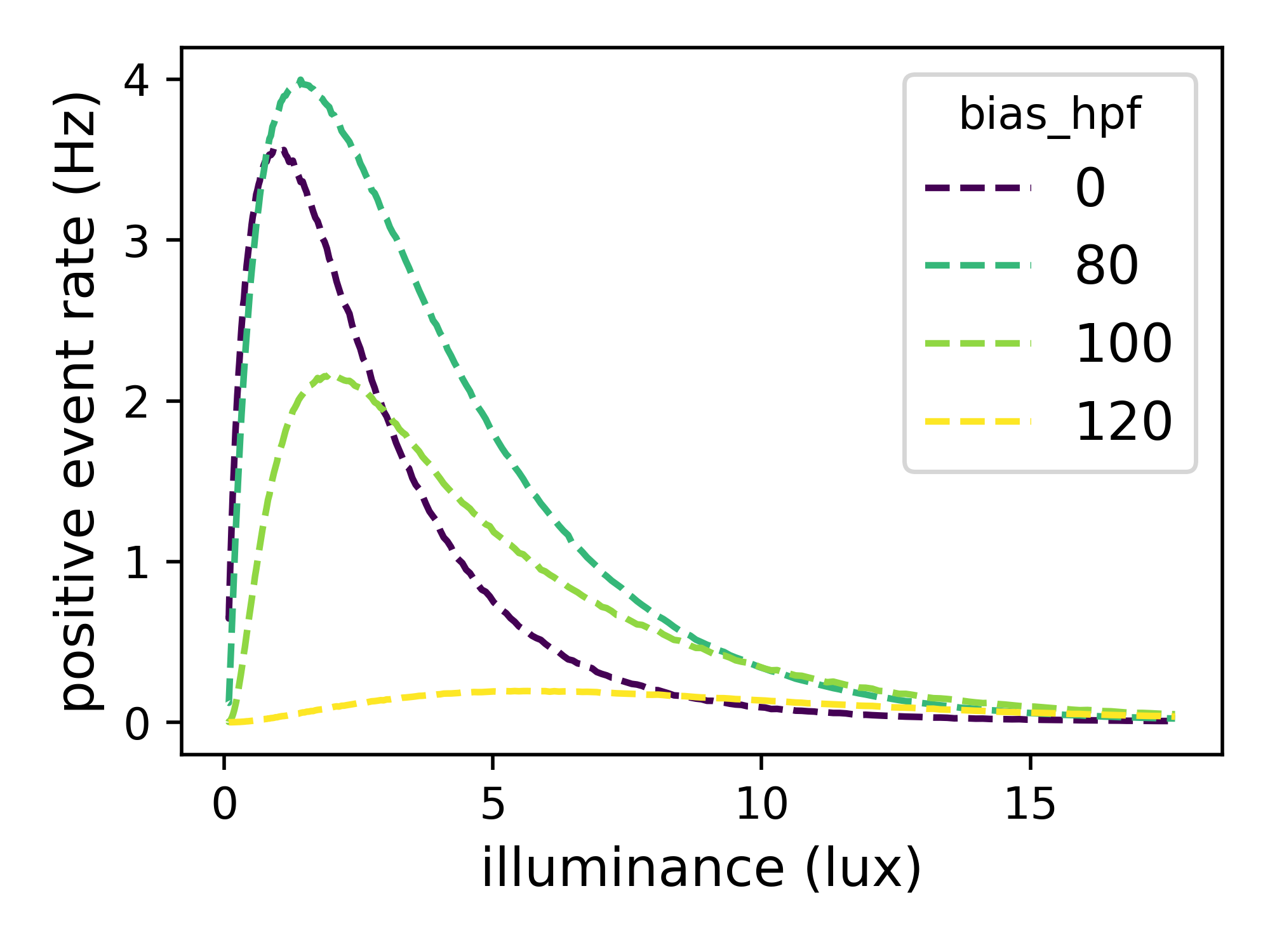}  
\caption{Experimentally captured noise event rate vs. illuminance levels using various sensor high-pass filter bias values which is called ``bias\_hpf'' in Prophesee Metavision SDK. The correlation between illuminance and event rate changes with different bias\_hpf values. } \label{fig-hpf}
\end{figure}

\renewcommand{\tablename}{Supplementary Table}
\setcounter{table}{0}

\begin{table}
\centering
\begin{tabular}{ p{3.2cm}p{1cm}p{1cm}p{1cm}c p{1cm}p{1cm}p{1cm} }
\hline
\multirow{2}{*}{Aggregation duration} & \multicolumn{3}{c}{In-distribution test} & & \multicolumn{3}{c}{Out-of-distribution test} \\
\cline{2-4} \cline{6-8}
                    & PSNR & SSIM & LPIPS && PSNR & SSIM & LPIPS \\
\hline
0.1 s               & 23.8 & 0.714& 0.391 && 18.4 & 0.513& 0.586 \\
0.25 s              & 24.4 & 0.721& 0.377 && 18.8 & 0.522& 0.564 \\
0.5 s               & 24.7 & 0.738& 0.358 && 19.1 & 0.545& 0.537 \\
1 s                 & 25.0 & 0.742& 0.349 && 19.3 & 0.552& 0.509 \\
2 s                 & 25.4 & 0.765& 0.324 && 19.5 & 0.598& 0.473 \\
\hline
\end{tabular}
\caption{Noise2Image performance using noise event counts aggregated over different time windows. For each aggregation duration, the Noise2Image model is trained and evaluated using the experimental data with identical training setting.}
\label{table-time}
\end{table}

\end{document}